\documentclass{article}
\usepackage{ijcai18}

\usepackage{times}
\usepackage{xcolor}
\usepackage{soul}
\usepackage[utf8]{inputenc}
\usepackage[small]{caption}
\usepackage{setspace}

\usepackage{epsfig}
\usepackage{graphicx}
\usepackage{amsmath}
\usepackage{amssymb}
\usepackage{amsfonts}
\usepackage[ruled,vlined]{algorithm2e}
\usepackage{multirow} 
\usepackage{mathrsfs}
\usepackage{subfigure}
\usepackage{url}
\newtheorem{definition}{Definition} 

\newcommand{\tr}{\operatornamewithlimits{tr}}
\newcommand{\HSIC}{\operatornamewithlimits{HSIC}}

\title{Ensemble Soft-Margin Softmax Loss for Image Classification }

\author{
Xiaobo Wang$^{1,2,*}$,
Shifeng Zhang$^{1,2,}$\thanks{These authors contributed equally to this work.},
Zhen Lei$^{1,2,}$\thanks{Corresponding author},
Si Liu$^3$,
Xiaojie Guo$^4$,
Stan Z. Li$^{5,1,2}$
\\
$^1$ CBSR\&NLPR, Institute of Automation, Chinese Academy of Sciences, Beijing, China \\
$^2$ University of Chinese Academy of Sciences, Beijing, China.\\
$^3$ Beijing Key Laboratory of Digital Media, School of Computer Science and Engineering, Beihang University.  \\
$^4$ School of Computer Software, Tianjin University, Tianjin, China.  \\
$^5$ Faculty of Information Technology, Macau University of Science and Technology, Macau, China.  \\
\{xiaobo.wang, shifeng.zhang, zlei, szli\}@nlpr.ia.ac.cn, fifthzombiesi@gmail.com, xj.max.guo@gmail.com
}

\begin{document}

\maketitle

\begin{abstract}
   Softmax loss is arguably one of the most popular losses to train CNN models for image classification. However, recent works have exposed its limitation on feature discriminability. This paper casts a new viewpoint on the weakness of softmax loss. On the one hand, the CNN features learned using the softmax loss are often inadequately discriminative. We hence introduce a soft-margin softmax function to explicitly encourage the discrimination between different classes. On the other hand, the learned classifier of softmax loss is weak. We propose to assemble multiple these weak classifiers to a strong one, inspired by the recognition that the diversity among weak classifiers is critical to a good ensemble. To achieve the diversity, we adopt the Hilbert-Schmidt Independence Criterion (HSIC). Considering these two aspects in one framework, we design a novel loss, named as Ensemble soft-Margin Softmax (EM-Softmax). Extensive experiments on benchmark datasets are conducted to show the superiority of our design over the baseline softmax loss and several state-of-the-art alternatives.
\end{abstract}

\section{Introduction}
Image classification is a fundamental yet still challenging task in machine learning and computer vision. Over the past years, deep \textit{Convolutional Neural Networks} (CNNs) have greatly boosted the performance of a series of image classification tasks, like object classification \cite{Alexnet,Resnet,L-softmax}, face verification \cite{Center,Range,SphereFace,NormFace} and hand-written digit recognition \cite{Maxout,NiN}, \textit{etc}. Deep networks naturally integrate low/mid/high-level features and classifiers in an end-to-end multi-layer fashion. Wherein each layer mainly consists of convolution, pooling and non-linear activation, leading CNNs to the strong visual representation ability as well as their current significant positions.

To train a deep model, the loss functions, such as (square) hinge loss, contrastive loss, triplet loss and softmax loss, \textit{etc.}, are usually equipped. Among them, the softmax loss is arguably the most popular one \cite{L-softmax}, which consists of three components, including the \textbf{last fully connected layer}, the \textbf{softmax function}, and the \textbf{cross-entropy loss}\footnote{The details of each component will be described in section \ref{softmaxoriginal}.} . It is widely adopted by many CNNs \cite{Alexnet,VGG,Resnet} due to its simplicity and clear probabilistic interpretation. However, the works \cite{L-softmax,Center,Range} have shown that the original softmax loss is inadequate due to the lack of encouraging the discriminability of CNN features. Recently, a renewed trend is to design more effective losses to enhance the performance. But this is non-trivial because a new designed loss usually should be easily optimized by the \textit{Stochastic Gradient Descent} (SGD) \cite{Pooling}.

To improve the softmax loss, existing works can be mainly categorized into two groups. One group \textit{tries to refine the cross-entropy loss of softmax loss}. Sun \textit{et al.} \cite{DeepID} trained the CNNs with the combination of softmax loss and contrastive loss, but the pairs of training samples are difficult to select. Schroff \textit{et al.} \cite{Facenet} used the triplet loss to minimize the distance between an anchor sample and a positive sample (of the same identity), as well as maximize the distance between the anchor and a negative sample (of different identities). However, requiring a multiple of three training samples as input makes it inefficient. Tang \textit{et al.} \cite{Hinge} replaced the cross-entropy loss with the hinge loss, while Liu \textit{et al.} \cite{Cosine} employed a congenerous cosine loss to enlarge the inter-class distinction as well as alleviate the inner-class variance. Its drawback is that these two losses are frequently unstable. Recently, Wen \textit{et al.} \cite{Center} introduced a center loss together with the softmax loss. Zhang \textit{et al.} \cite{Range} proposed a range loss to handle the case of the long tail distribution of data. Both of them have achieved promising results on face verification task. However, the objective of the open-set face verification (\textit{i.e.}, mainly to learn discriminative features) is different from that of the closed-set image classification (\textit{i.e.}, simultaneously to learn discriminative features and a strong classifier). The other group is to \textit{reformulate the softmax function of softmax loss}.Liu \textit{et al.} \cite{L-softmax,SphereFace} enlarged the margin of the softmax function to encourage the discriminability of features and further extended it to the face verification task. Wang \textit{et al.} \cite{NormFace} developed a normalized softmax function to learn discriminative features. However, for the \textit{last fully connected layer}\footnote{For convenience, we denote it as softmax classifier.}\textit{of softmax loss}, few works have considered. The fully convolutional networks \cite{R-FCN} and the global average pooling \cite{NiN,CAM} aim to modify the fully connected layers of DNNs, they are not applicable to the softmax classifier. In fact, for deep image classification, the softmax classifier is of utmost importance.

Since feature extracting and classifier learning in CNNs are in an end-to-end framework, in this paper, we argue that the weakness of softmax loss mainly comes from two aspects. One is that the extracted features are not discriminative. The other one is that the learned classifier is not strong. To address the above issues, we introduce a simple yet effective soft-margin softmax function to explicitly emphasize the feature discriminability, and adopt a novel ensemble strategy to learn a strong softmax classifier. For clarity, our main contributions are summarized as follows:
\begin{itemize}
\item{We cast a new viewpoint on the weakness of the original softmax loss. \textit{i.e.}, the extracted CNN features are insufficiently discriminative and the learned classifier is weak for deep image classification.}
\item{We design a soft-margin softmax function to encourage the feature discriminability and attempt to assemble the weak classifiers of softmax loss by employing the Hilbert-Schmidt Independence Criterion (HSIC).}
\item{We conduct experiments on the datasets of MNIST, CIFAR10/CIFAR10+, CIFAR100/CIFAR100+, and ImageNet32 \cite{Imagenet32}, which reveal the effectiveness of the proposed method. }
\end{itemize}

\section{Preliminary Knowledge}
\subsection{Softmax Loss}\label{softmaxoriginal}
Assume that the output of a single image through deep convolution neural networks is $\mathbf{x}$ (\textit{i.e.}, CNN features), where $\mathbf{x} \in \mathbb{R}^d$, $d$ is the feature dimension. Given a mini-batch of labeled images, their outputs are $\{\mathbf{x}_1,\mathbf{x}_2,\dots,\mathbf{x}_n\}$. The corresponding labels are $\{y_1,y_2,\dots,y_n\}$, where $y_i \in \{1,2,\dots,K\}$ is the class indicator, and $K$ is the number of classes. Similar to the work \cite{L-softmax}, we define the complete softmax loss as the pipeline combination of the last fully connected layer, the softmax function and the cross-entropy loss. The {\textbf{last fully connected layer}} transforms the feature $\mathbf{x}$ into a primary score $\mathbf{z}= [z_1,z_2,\dots,z_K]^T \in \mathbb{R}^K$ through multiple parameters $\mathbf{W}=[\mathbf{w}_1,\mathbf{w}_2,\dots,\mathbf{w}_K] \in \mathbb{R}^{d\times K}$, which is formulated as: $z_k = \mathbf{w}_k^T \mathbf{x} = \mathbf{x}^T\mathbf{w}_k$. Generally speaking, the parameter $\mathbf{w}_k$ can be regarded as the linear classifier of class $k$. Then, the \textbf{softmax function} is applied to transform the primary score $z_k$ into a new predicted class score as: $s_{k} = \frac{e^{z_k}}{e^{z_k}+\sum_{t\neq k}^Ke^{z_t}}$. Finally, the \textbf{cross-entropy loss} $\mathcal{L}_{\rm{S}} = - \log(s_y)$ is employed.

\subsection{Hilbert-Schmidt Independence Criterion} \label{HSICdefinition}
The \textit{Hilbert-Schmidt Independence Criterion} (HSIC) was proposed in \cite{HSIC} to measure the (in)dependence of two random variables $\mathcal{X}$ and $\mathcal{Y}$.
\begin{definition} \label{EHSIC}
\rm{\textbf{(HSIC)}} Consider $n$ independent observations drawn from $p_{\mathbf{xy}}$, $\mathcal{Z} := \{(\mathbf{x}_1, \mathbf{y}_1),\dots,(\mathbf{x}_n, \mathbf{y}_n)\} \subseteq
\mathcal{X} \times \mathcal{Y}$, an empirical estimator of HSIC($\mathcal{Z},\mathcal{F}, \mathcal{G}$), is given by:
\begin{equation}\label{HSIC}
\HSIC (\mathcal{Z},\mathcal{F}, \mathcal{G}) = (n-1)^{-2} \tr (\mathbf{K}_1\mathbf{H}\mathbf{K}_2\mathbf{H}),
\end{equation}
where $\mathbf{K}_1$ and $\mathbf{K}_2$ are the Gram matrices with $k_{1,ij} = k_1(\mathbf{x}_i, \mathbf{x}_j)$, $k_{2,ij} = k_2(\mathbf{y}_i, \mathbf{y}_j)$. $k_1(\mathbf{x}_i, \mathbf{x}_j)$ and $k_2(\mathbf{y}_i, \mathbf{y}_j)$ are the kernel functions defined in space $\mathcal{F}$ and $\mathcal{G}$, respectively. $\mathbf{H}=\mathbf{I}-n^{-1}\mathbf{1}\mathbf{1}^T$ centers the Gram matrix to have zero mean.\end{definition}
\noindent Note that, according to Eq. (\ref{HSIC}), to maximize the independence between two random variables $\mathcal{X}$ and $\mathcal{Y}$, the empirical estimate of HSIC, \textit{i.e.}, $\tr(\mathbf{K}_1\mathbf{H}\mathbf{K}_2\mathbf{H})$ should be minimized.


\section{Problem Formulation}
Inspired by the recent works \cite{L-softmax,Center,Range}, which argue that the original softmax loss is inadequate due to its non-discriminative features. They either reformulate the softmax function into a new desired one (\textit{e.g.}, L-softmax \cite{L-softmax} \textit{etc.}) or add additional constraints to refine the original softmax loss (\textit{e.g.}, contrastive loss \cite{DeepID} and center loss \cite{Center} \textit{etc.}). Here, we follow this argument but cast a new viewpoint on the weakness, say the extracted features are not discriminative meanwhile the learned classifier is not strong.

\subsection{Soft-Margin Softmax Function}
To enhance the discriminability of CNN features, we design a new soft-margin softmax function to enlarge the margin between different classes. We first give a simple example to describe our intuition. Consider the binary classification and we have a sample $\mathbf{x}$ from class 1. The original softmax loss is to enforce $\mathbf{w}_1^T\mathbf{x} > \mathbf{w}_2^T\mathbf{x}$ ( \textit{i.e.}, $\|\mathbf{w}_1\|\|\mathbf{x}\|\cos(\theta_1)>\|\mathbf{w}_2\|\|\mathbf{x}\|\cos(\theta_2)$) to classify $\mathbf{x}$ correctly. To make this objective more rigorous, the work L-Softmax \cite{L-softmax} introduced an angular margin:
\begin{equation}
\begin{aligned}
\|\mathbf{w}_1\|\|\mathbf{x}\|\cos(\theta_1) \geq \|\mathbf{w}_1\|\|\mathbf{x}\|\cos(m\theta_1) >\|\mathbf{w}_2\|\|\mathbf{x}\|\cos(\theta_2),
\end{aligned}
\end{equation}
and used the intermediate value $\|\mathbf{w}_1\|\|\mathbf{x}\|\cos(m\theta_1)$ to replace the original $\|\mathbf{w}_1\|\|\mathbf{x}\|\cos(\theta_1)$ during training. In that way, the discrmination between class 1 and class 2 is explicitly emphasized. However, to make $\cos(m\theta_1)$ derivable, $m$ should be a positive integer. In other words, the angular margin cannot go through all possible angles and is a \textit{hard} one. Moreover, the forward and backward computation are complex due to the angular margin involved. To address these issues, inspired by the works \cite{DeepID,SM-Softmax,Bell}, we here introduce a \textit{soft} distance margin and simply let
\begin{equation} \label{softmargin}
\mathbf{w}_1^T\mathbf{x}\geq \mathbf{w}_1^T\mathbf{x}-m > \mathbf{w}_2^T\mathbf{x},
\end{equation}
where $m$ is a non-negative real number and is a distance margin. In training, we employ $\mathbf{w}_1^T\mathbf{x}-m$ to replace $\mathbf{w}_1^T\mathbf{x}$, thus our multi-class soft-margin softmax function can be defined as: $s_{i} =\frac{e^{\mathbf{w}_{y}^T\mathbf{x}_i-m}}{e^{\mathbf{w}_{y}^T\mathbf{x}_i-m}+\sum_{k\neq y}^Ke^{\mathbf{w}_{k}^T\mathbf{x}_i}}.$ Consequently, the soft-Margin Softmax (M-Softmax) loss is formulated as:
\begin{equation}\label{M-Softmax}
\begin{aligned}
\mathcal{L}_{\rm{M}} = -\log\frac{e^{\mathbf{w}_{y}^T\mathbf{x}_i-m}}{e^{\mathbf{w}_{y}^T\mathbf{x}_i-m}+\sum_{k\neq y}^Ke^{\mathbf{w}_{k}^T\mathbf{x}_i}}.
\end{aligned}
\end{equation}
Obviously, when $m$ is set to zero, the designed M-Softmax loss Eq. (\ref{M-Softmax}) becomes identical to the original softmax loss.

\subsection{Diversity Regularized Ensemble Strategy} \label{NE-Softmax}
Though learning discriminative features may result in better classifier as these two components highly depend on each other, the classifier may not be strong enough without explicitly encouragement. To learn a strong one, as indicted in \cite{ERM}, a combination of various classifiers can improve predictions. Thus we adopt the ensemble strategy. Prior to formulating our ensemble strategy, we revisit that the most popular way to train an ensemble in deep learning is arguably the dropout \cite{Dropout}. The idea behind dropout is to train an ensemble of DNNs by randomly dropping the activations and average the results of the whole ensemble instead of training a single DNN. However, in the last fully connected layer of softmax loss, dropout is usually not permitted because it will lose the useful label information, especially with the limited training samples. Therefore, we need to design a new manner to assemble weak classifiers.

Without loss of generality, we take two weak softmax classifiers $\mathbf{W}_v=[\mathbf{w}_1^v,\mathbf{w}_2^v,\dots,\mathbf{w}_K^v]\in \mathbb{R}^{d \times K}$ and $\mathbf{W}_u=[\mathbf{w}_1^u,\mathbf{w}_2^u,\dots,\mathbf{w}_K^u]\in \mathbb{R}^{d \times K}$ as an example to illustrate the main idea. Specifically, it has been well-recognized that the diversity of weak classifiers is of utmost importance to a good ensemble \cite{ERM,DRM}. Here, we exploit the diverse/complementary information across different weak classifiers by enforcing them to be independent. High independence of two weak classifiers $\mathbf{W}_v$ and $\mathbf{W}_u$ means high diversity of them. Classical independence criteria like the Spearmans rho and Kendalls tau \cite{Independence}, can only exploit linear dependence. The recent exclusivity regularized term \cite{ERM,ECMSC} and ensemble pruning \cite{DRM} may be good candidates for classifier ensemble, but both of them are difficult to differentiate. Therefore, these methods are not suitable for assembling the weak softmax classifiers.

In this paper, we employ the Hilbert-Schmidt Independence Criterion (HSIC) to measure the independence (\textit{i.e.}, diversity) of weak classifiers, mainly for two reasons. One is that the HSIC measures the dependence by mapping variables into a Reproducing Kernel Hilbert Space (RKHS), such that the nonlinear dependence can be addressed. The other one is that the HSIC is computational efficient. The empirical HSIC in Eq. (\ref{HSIC}) turns out to be the trace of product of weak classifiers, which can be easily optimized by the typical SGD. Based on the above analysis, we naturally minimize the following constraint according to Eq. (\ref{HSIC}):
\begin{equation}\label{Zhang}
\HSIC(\mathbf{W}_v,\mathbf{W}_u) = (K-1)^{-2}\tr(\mathbf{K}_v\mathbf{H}\mathbf{K}_u\mathbf{H}).
\end{equation}
For simplicity, we adopt the inner product kernel for the proposed HSIC, say $\mathbf{K} = \mathbf{W}^T\mathbf{W}$ for both $\mathbf{W}_v$ and $\mathbf{W}_u$. Considering the multiple ensemble settings and ignoring the scaling factor $(K-1)^{-2}$ of HSIC for notational convenience, leads to the following equation:
\begin{equation}\label{HSICfinal}
\begin{aligned}
&\sum_{u=1;u\neq v}^V \HSIC(\mathbf{W}_v,\mathbf{W}_u) = \sum_{u=1;u\neq v}^V \tr(\mathbf{K}_v\mathbf{H}\mathbf{K}_u\mathbf{H}) \\
&= \sum_{u=1;u\neq v}^V \tr(\mathbf{W}_v\mathbf{H}\mathbf{W}_u^T\mathbf{W}_u\mathbf{H}\mathbf{W}_v^T) = \tr(\mathbf{W}_v\mathbf{K}^v\mathbf{W}_v^T),
\end{aligned}
\end{equation}
where $\mathbf{K}^v = \sum_{u=1;u\neq v}^V \mathbf{H}\mathbf{W}_u^T\mathbf{W}_u\mathbf{H}$, $\mathbf{H}$ is the centered matrix defined in section \ref{HSICdefinition}, and $\mathbf{H}^T=\mathbf{H}$. However, according to the formulation Eq. (\ref{HSICfinal}), we can see that the HSIC constraint is value-aware, the diversity is determined by the value of weak classifiers. If the magnitude of different weak classifiers is quite large, the diversity may not be well handled. To avoid the scale issue, we use the normalized weak classifiers to compute the diversity. In other words, if not specific, the weak classifiers $\mathbf{W}_v$, where $v \in \{1,2,\dots,V\}$ are normalized in Eq. (\ref{HSICfinal}). Merging the diversity constraint into softmax loss, leads to Ensemble Softmax (E-Softmax) loss as:
\begin{equation}\label{DE-Softmax}
\begin{aligned}
\mathcal{L}_{\rm{E}} =  \sum_{v=1}^V \Big[- \log \frac{e^{\mathbf{x}_i^T\mathbf{w}_y^v}}{\sum_{k=1}^Ke^{\mathbf{x}_i^T\mathbf{w}_k^v}}  + \lambda \tr(\mathbf{W}_v\mathbf{K}^v\mathbf{W}_v^T) \Big],
\end{aligned}
\end{equation}
where $\lambda$ is a hyperparameter to balance the importance of diversity. The backward propagation of $\mathbf{W}_v$ is computed as  $\frac{\partial{\mathcal{L}_{\rm{E}}}}{\partial{\mathbf{W}_v}} = \frac{\partial{\mathcal{L}_{\rm{S}}}}{\partial{\mathbf{W}_v}} + \lambda\mathbf{W}_v\mathbf{K}^v$. Clearly, the update of the weak classifier $\mathbf{W}_v$ is co-determined by the initializations and other weak classifiers (\textit{i.e.}, $\mathbf{K}^v$ is computed based on other classifiers). This means that the diversity of different weak classifiers will be explicitly enhanced.

Since feature extracting and classifier learning is an end-to-end framework, we prefer to simultaneously harness them. Now, putting all concerns, say Eqs. (\ref{M-Softmax}) and (\ref{HSICfinal}), together results in our final Ensemble soft-Margin Softmax (EM-Softmax) loss $\mathcal{L}_{\rm{EM}}$ as follows:
\begin{equation}\label{EM-Softmax}
\begin{aligned}
 \sum_{v=1}^V \Big[- \log \frac{e^{\mathbf{x}_i^T\mathbf{w}_y^v-m}}{e^{\mathbf{x}_i^T\mathbf{w}_y^v-m}+\sum_{k\neq y}^Ke^{\mathbf{x}_i^T\mathbf{w}_k^v}} + \lambda \tr(\mathbf{W}_v\mathbf{K}^v\mathbf{W}_v^T) \Big].
\end{aligned}
\end{equation}

\subsection{Optimization}\label{Optimization}
In this part, we show that the proposed EM-Softmax loss is trainable and can be easily optimized by the typical SGD. Specifically, we implement the CNNs using the well-known Caffe \cite{caffe} library and use the chain rule to compute the partial derivative of each $\mathbf{W}_v$ and the feature $\mathbf{x}$ as:
\begin{equation}
\begin{aligned}
\frac{\partial{\mathcal{L}_{\rm{EM}}}}{\partial{\mathbf{W}_v}} &=\frac{\partial{\mathcal{L}_{\rm{S}}}}{\partial{\mathbf{s}_v}}\frac{\partial{\mathbf{s}_v}}{\partial{\mathbf{z}_v}}\frac{\partial{\mathbf{z}_v}}{\partial{\mathbf{W}_v}} + \lambda \mathbf{W}_v\mathbf{K}^v, \\
\frac{\partial{\mathcal{L}_{\rm{EM}}}}{\partial{\mathbf{x}}} &= \sum_{v=1}^V \frac{\partial{\mathcal{L}_{\rm{S}}}}{\partial{\mathbf{s}_v}}\frac{\partial{\mathbf{s}_v}}{\partial{\mathbf{z}_v}}\frac{\partial{\mathbf{z}_v}}{\partial{\mathbf{x}}} ,
\end{aligned}
\end{equation}
where the computation forms of $\frac{\partial{\mathcal{L}_{\rm{S}}}}{\partial{\mathbf{s}_v}}$, $\frac{\partial{\mathbf{s}_v}}{\partial{\mathbf{z}_v}}$ $\frac{\partial{\mathbf{z}_v}}{\partial{\mathbf{x}}}$, $\frac{\partial{\mathbf{z}_v}}{\partial{\mathbf{W}_v}}$ are the same as the original softmax loss.

\section{Experiments}
\subsection{Dataset Description}\label{dataset}

\noindent \textbf{MNIST} \cite{MNIST}: The MNIST is a dataset of handwritten digits (from 0 to 9) composed of $28\times 28$ pixel gray scale images. There are 60, 000 training images and 10, 000 test images. We scaled the pixel values to the $[0, 1]$ range before inputting to our neural network.

\noindent \textbf{CIFAR10}/\textbf{CIFAR10+} \cite{CIFAR10}: The CIFAR10 contains 10 classes, each with 5, 000 training samples and 1, 000 test samples. We first compare EM-Softmax loss with others under no data augmentation setup. For the data augmentation, we follow the standard technique in \cite{Data4,L-softmax} for training, that is, 4 pixels are padded on each side, and a $32\times 32$ crop is randomly sampled from the padded image or its horizontal flip. In testing, we only evaluate the single view of the original $32\times 32$ image. In addition, before inputting the images to the network, we subtract the per-pixel mean computed over the training set from each image.

\noindent \textbf{CIFAR100/CIFAR100+} \cite{CIFAR10}: We also evaluate the performance of the proposed EM-Softmax loss on CIFAR100 dataset. The CIFAR100 dataset is the same size and format as the CIFAR10 dataset, except it has 100 classes containing 600 images each. There are 500 training images and 100 testing images per class. For the data augmentation set CIFAR100+, similarly, we follow the same technique provided in \cite{Data4,L-softmax}.

\noindent \textbf{ImageNet32} \cite{Imagenet32}: The ImageNet32 is a downsampled version of the ImageNet 2012 challenge dataset, which contains exactly the same number of images as the original ImageNet, \textit{i.e.}, 1281, 167 training images and 50, 000 validation images for 1, 000 classes. All images are downsampled to $32\times32$. Similarly, we subtract the per-pixel mean computed over the downsampled training set from each image before feeding them into the network.

\begin{table*}[t]
	\begin{center}
\scalebox{1.0}{
		\begin{tabular}{|c|c|c|c|c|c|c| }
		   \hline
		   &Method & \ \ MNIST \ \ & \ CIFAR10 \ & \ \ CIFAR10+ \ \ & \ CIFAR100 \ & \ CIFAR100+ \ \\
		   \hline
           \hline
           \multirow{5}*{Compared}
           & HingeLoss \cite{Hinge}         & 99.53* & 90.09* & 93.31* & 67.10* & 68.48 \\
           & CenterLoss \cite{Center}       & 99.41 & 91.65 & 93.82 & 69.23 & 70.97 \\
           & A-Softmax \cite{SphereFace}    & 99.66 & 91.72 & 93.98 & 70.87 & 72.23 \\
           & N-Softmax \cite{NormFace}      & 99.48 & 91.46 & 93.90 & 70.49 & 71.85 \\
           & L-Softmax \cite{L-softmax}     & 99.69* & 92.42* & 94.08* & 70.47* & 71.96 \\
           \hline
           Baseline& Softmax & 99.60* & 90.95* & 93.50* & 67.26* & 69.15 \\
           \hline
           \multirow{3}*{Our}
           &M-Softmax  & 99.70 & 92.50 & 94.27 & 70.72 & 72.54 \\
           &E-Softmax  & 99.69 & 92.38 & 93.92 & 70.34 & 71.33 \\
           \cline{2-7}
           &EM-Softmax  & \textbf{99.73} & \textbf{93.31} & \textbf{95.02} & \textbf{72.21} &  \textbf{75.69} \\
           \hline
        \end{tabular} }
	 \end{center}		
\caption{Recognition accuracy rate (\%) on \underline{MNIST}, \underline{CIFAR10/CIFAR10+} and \underline{CIFAR100/CIFAR100+} datasets.}
\label{Benchmark}
\end{table*}

\subsection{Compared Methods}
We compare our EM-Softmax loss with recently proposed state-of-the-art alternatives, including the baseline softmax loss (Softmax), the margin-based hinge loss (HingeLoss \cite{Hinge}), the combination of softmax loss and center loss (CenterLoss \cite{Center}), the large-margin softmax loss (L-Softmax \cite{L-softmax}), the angular margin softmax loss (A-Softmax \cite{SphereFace}) and the normalized features softmax loss (N-Softmax \cite{NormFace}). The source codes of Softmax and HingeLoss are provided in Caffe community. For other compared methods, their source codes can be downloaded from the github or from authors' webpages. For fair comparison, the experimental results are cropped from the paper \cite{L-softmax} (indicated as *) or obtained by trying our best to tune their corresponding hyperparameters. Moreover, to verify the gain of our soft margin and ensemble strategy, we also report the results of the M-Softmax loss Eq. (\ref{M-Softmax}) and the E-Softmax loss Eq. (\ref{DE-Softmax}).

\subsection{Implementation Details}\label{Architecture1}
In this section, we give the major implementation details on the baseline works and training/testing settings. The important statistics are provided as follows:

\noindent \textbf{Baseline works}. To verify the universality of EM-Softmax, we choose the work \cite{L-softmax} as the baseline. We strictly follow all
experimental settings in \cite{L-softmax}, including the CNN architectures (LiuNet\footnote{The detailed CNNs for each dataset can be found at \url{https://github.com/wy1iu/LargeMargin_Softmax_Loss}.}), the datasets, the pre-processing methods and the evaluation criteria.

\noindent \textbf{Training}. The proposed EM-Softmax loss is appended after the feature layer, \textit{i.e.}, the second last inner-product layer. We start with a learning rate of 0.1, use a weight decay of 0.0005 and momentum of 0.9. For MNIST, the learning rate is divided by 10 at 8k and 14k iterations. For CIFAR10/CIFAR10+, the learning rate is also divided by 10 at 8k and 14k iterations. For CIFAR100/CIFAR100+, the learning rate is divided by 10 at 12k and 15k iterations. For all these three datasets, the training eventually terminates at 20k iterations. For ImageNet32, the learning rate is divided by 10 at 15k, 25k and 35k iterations, and the maximal iteration is 40k. The accuracy on validation set is reported.

\noindent \textbf{Testing}. At testing stage, we simply construct the final ensemble classifier by averaging weak classifiers as: $\mathbf{W}=\frac{1}{V}\sum_{v=1}^V \mathbf{W}_v$. Finally, $\mathbf{W}$ is the learned strong classifier and we use it with the discriminative feature $\mathbf{x}$ to predict labels.

\begin{figure} \centering
	\includegraphics[width=0.495\columnwidth]{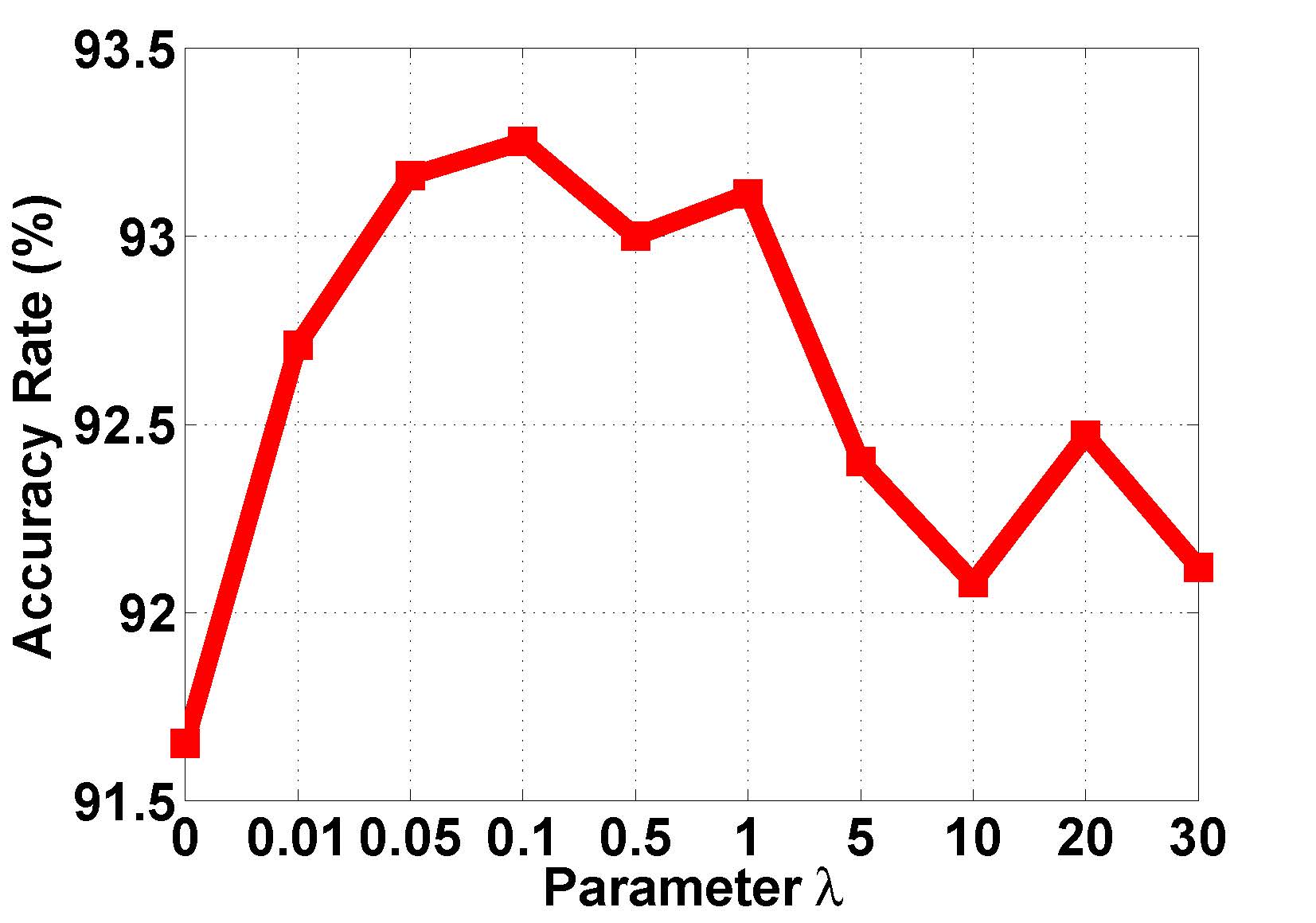}
 \includegraphics[width=0.495\columnwidth]{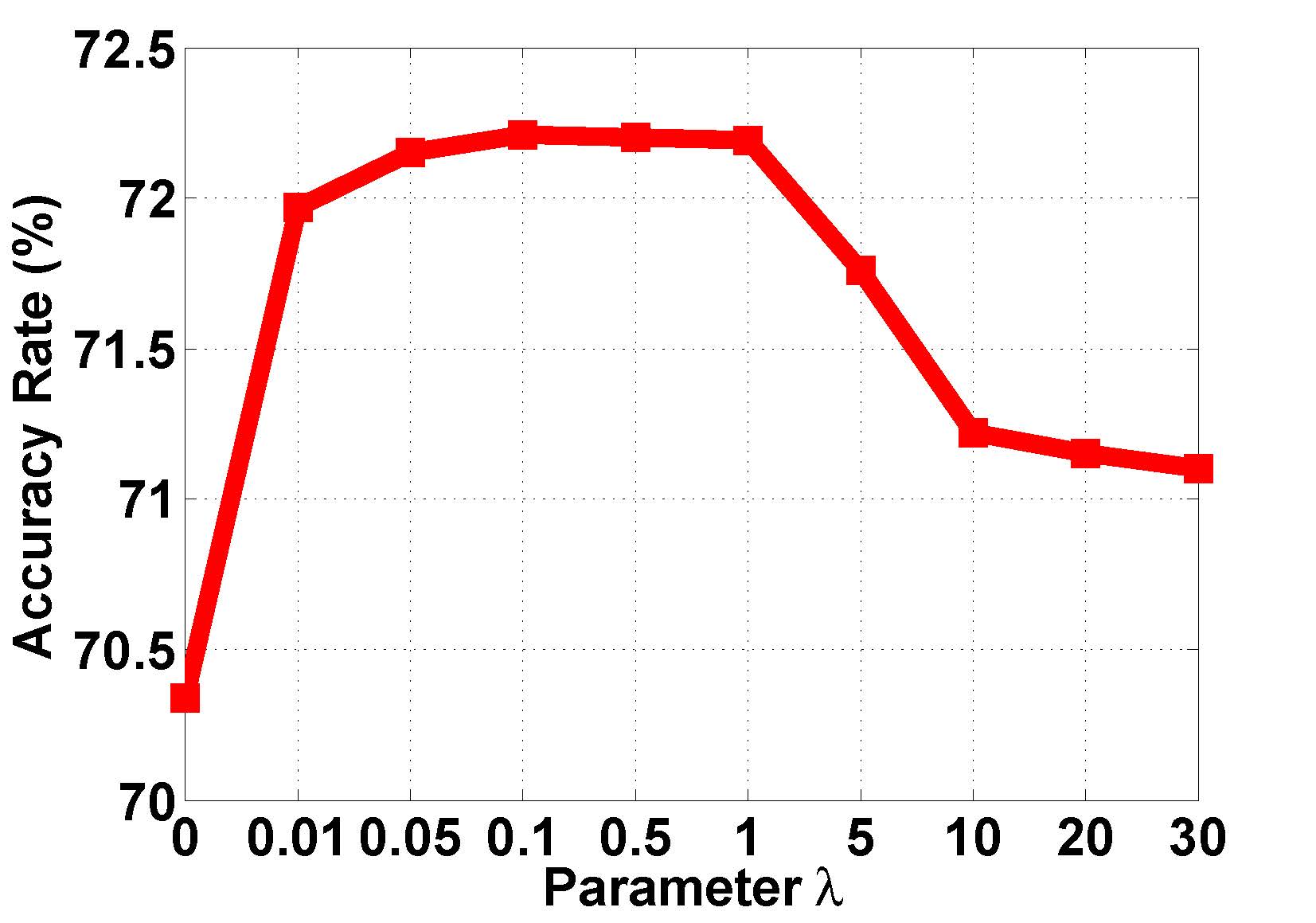}
	\caption{\textbf{From Left to Right}: The effects parameter $\lambda$ of EM-Softmax loss on CIFAR10 and CIFAR100, respectively. The results are obtained by tuning different $\lambda$ with $V=2$.}
	\label{ParameterEffect}
\end{figure}

\subsection{Accuracy vs HyperParameter} \label{Parameter}
\begin{figure} \centering
	\includegraphics[width=0.495\columnwidth]{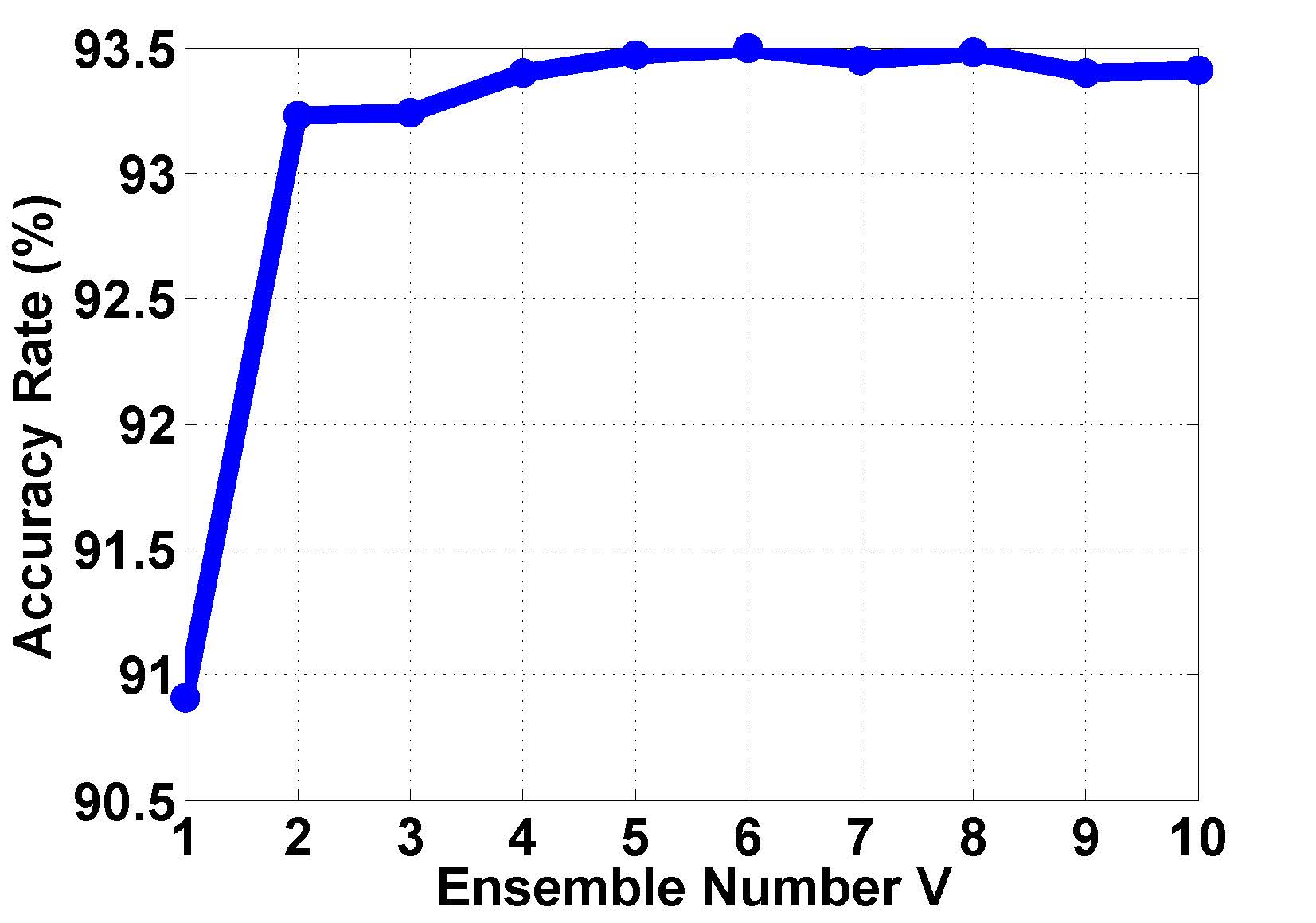}
	\includegraphics[width=0.495\columnwidth]{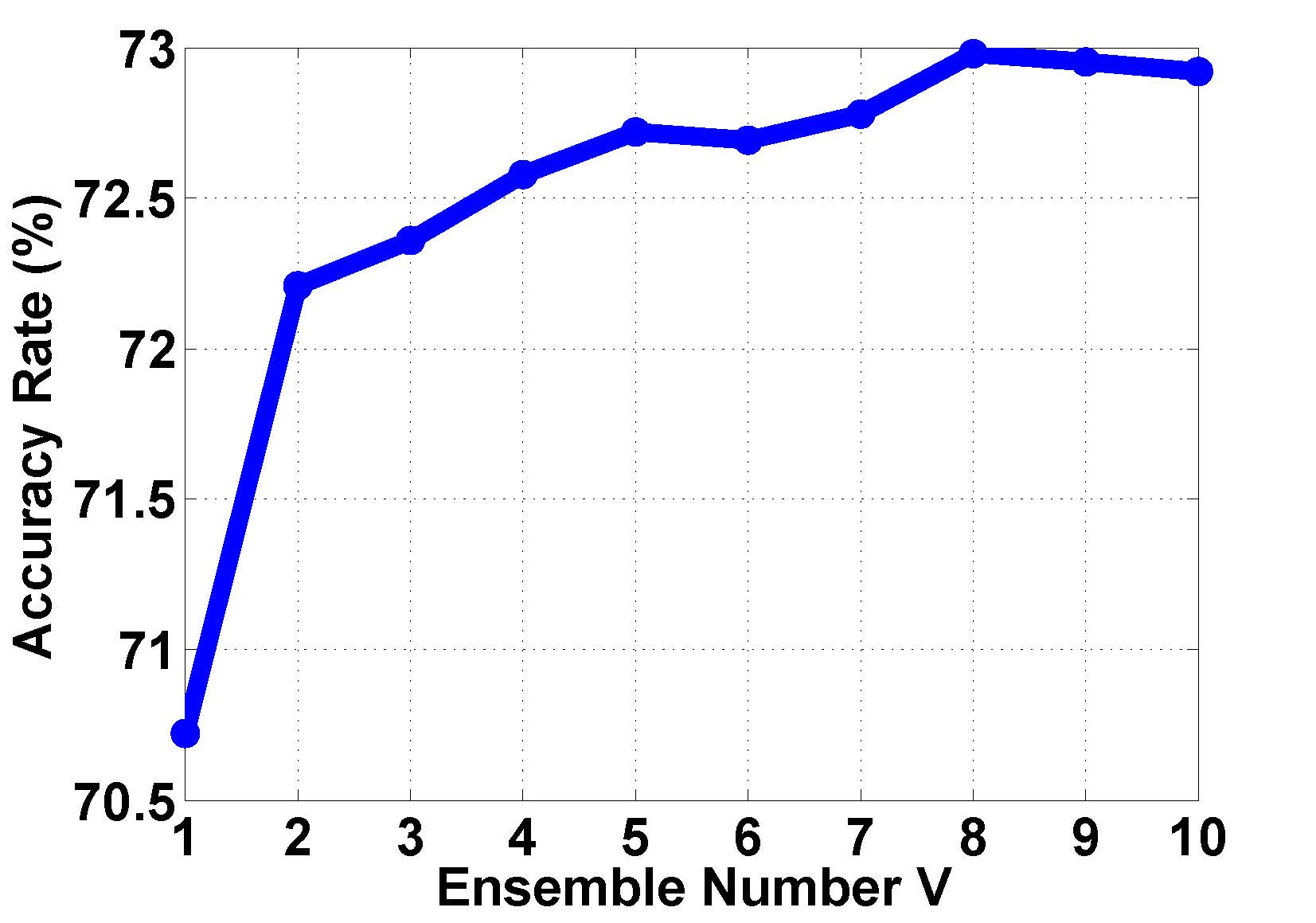}
	\caption{\textbf{From left to right}: The effects of ensemble number $V$ on CIFAR10 and CIFAR100, respectively. The results are obtained by tuning different ensemble number $V$ with $\lambda=0.1$.}
	\label{ParametervEffect}
\end{figure}

The soft-margin softmax function (\ref{M-Softmax}) involves one parameter $m$. Inspired by \cite{Bell}, we try a few different $m \in \{0.1, 0.5, 1, 5, 10\}$ and select the one that performs best. Regarding the diversity regularization (\ref{HSICfinal}), it involves the trade-off parameter $\lambda$ and the ensemble number $V$. In this part, we mainly report the sensitiveness of these two variables on CIFAR10 and CIFAR100. The subfigures of Figure \ref{ParameterEffect} displays the testing accuracy rate vs. parameter $\lambda$ plot of EM-Softmax loss. We set $V=2$ and vary $\lambda$ from 0 to 30 to learn different models. From the curves, we can observe that, as $\lambda$ grows, the accuracy grows gradually at the very beginning and changes slightly in a relatively large range. The experimental results are insensitive to $\lambda \in [0.01, \ 1.0]$. Too large $\lambda$ may hinder the performance because it will degenerate the focus of classification part in Eq. (\ref{EM-Softmax}). Moreover, it also reveals the effectiveness of the diversity regularization ($\lambda\neq 0$ vs. $\lambda=0$). The subfigures of Figure \ref{ParametervEffect} is the testing accuracy rate vs. ensemble number $V$ plot of EM-Softmax loss. We set $\lambda=0.1$ and vary $V$ from 1 to 10. From the curves, we can see that a single classifier ($V=1$) is weak for classification. Our EM-Softmax ($V\geq2$) benefits from the ensemble number of weak classifiers. But the improvement is slight when the ensemble number $V$ is big enough. The reason behind this may come from two aspects. One is that too many classifiers ensemble will bring too much redundant information, thus the improvement is finite. The other one is that the discriminative features help to promote weak classifiers, without needing assemble too many classifiers. Based on the above observations, we empirically suggest $V=2$ in practice to avoid the parameter explosion of weak classifiers.

\subsection{Classification Results on MNIST and CIFAR}
Table \ref{Benchmark} provides the quantitative comparison among all the competitors on MNIST and CIFAR datasets. The bold number in each column represents the best performance.

On MNIST dataset, it is well-known that this dataset is typical and easy for deep image classification, and all the competitors can achieve over 99\% accuracy rate. So the improvement of our EM-Softmax is not quite large. From the experimental results, we observe that A-Softmax \cite{SphereFace}, L-Softmax \cite{L-softmax}, the proposed EM-Softmax and its degenerations M-Softmax and E-Softmax outperform the other compared methods. Moreover, we have achieved a high accuracy rate 99.73\% on this dataset.

On CIFAR10/CIFAR10+ dataset, we can see that our EM-Softmax significantly boosts the performance, achieving at least 2\% improvement over the baseline Softmax. Considering all the competitors can achieve over 90\% accuracy rate on this dataset, the improvement is significant. To the soft distance margin M-Softmax, it is slightly better than the hard angular margin L-Softmax \cite{L-softmax} and A-Softmax \cite{SphereFace} because the soft margin can go through all possible desired ones. It is much better than Softmax because of the learned discriminative features. To the ensemble softmax E-Softmax, it is about 1\% higher than the baseline Softmax because of the assembled strong classifier. Our EM-Softmax absorbs the complementary merits from these two aspects (\textit{i.e.}, discriminative features and strong classifier).

On CIFAR100/CIFAR100+ dataset, it can be observed that most of competitors achieve relatively low performance. The major reason is that the large variation of subjects, color and textures and the fine-grained category involve in this dataset. Even so, our EM-Softmax still reaches significant improvements, at least 5\% higher than the baseline Softmax. Compared with the recent L-Softmax \cite{L-softmax} and A-Softmax \cite{SphereFace}, our EM-Softmax can achieve about 3\% improvement. Moreover, we can see a similar trend as that shown on CIFAR10/CIFAR10+ dataset, that is, the EM-Softmax loss is generally better than its degenerations M-Softmax and E-Softmax.

\begin{table}[t]
	\begin{center}
\scalebox{0.91}{
		\begin{tabular}{|c|c|c|c| }
		   \hline
		   &Method & Top-1 & Top-5  \\
		   \hline
           \hline
           \multirow{5}*{Compared}
           & HingeLoss \cite{Hinge}         & 46.52 & 71.56  \\
           & CenterLoss \cite{Center}       & 47.43 & 71.98 \\
           & A-Softmax \cite{SphereFace}    & 48.12 & 72.51 \\
           & N-Softmax \cite{NormFace}      & 47.52 & 72.06 \\
           & L-Softmax \cite{L-softmax}     & 47.85 & 72.63 \\
           \hline
           Baseline& Softmax & 46.89 & 71.94 \\
           \hline
           \multirow{3}*{Our}
           &M-Softmax  & 48.21  & 72.77 \\
           &E-Softmax  & 48.16 & 72.99 \\
           \cline{2-4}
           &EM-Softmax  & \textbf{49.22} & \textbf{74.22} \\
           \hline
        \end{tabular} }
	 \end{center}		
\caption{The top-1 and top-5 recognition accuracy rate (\%) on \underline{ImageNet32} validation set.}
\label{Imagenet}
\end{table}

\subsection{Classification Results on ImageNet}
We report the top-1 and top-5 validation accuracy rates on ImageNet32 in Table \ref{Imagenet}. From the numbers, we can see that the results exhibit the same phenomena that emerged on CIFAR dataset. In particular, the proposed EM-Softmax achieves a higher top-1 accuracy by 2.4\% and top-5 accuracy by 2.2\% in comparison with the baseline Softmax. The improvements are significant as the imagenet is very large and difficult for image classification, especially with such a smaller downsampled size (32$\times$32). Compared with other competitors, our EM-Softmax can achieve at least 1\% improvement. The results presented in Table \ref{Imagenet} also reveal that our EM-Softmax can benefits from the discriminative features (M-Softmax) and the strong classifier (E-Softmax).

\subsection{EM-Softmax vs. Model Averaging}
To validate the superiority of our weak classifiers ensemble strategy (\textit{i.e.}, EM-Softmax) over the simple model averaging, we conduct two kinds of model averaging experiments on both CIFAR10 and CIFAR100 datasets. One is the model averaging of the same architecture but with different numbers of filters (\textit{i.e.}, 48/48/96/192, 64/64/128/256 and 96/96/192/382)\footnote{64/64/128/256 denotes the number of filters in conv0.x, conv1.x, conv2.x and conv3.x, respectively.}. For convenience, we use CNN(48), CNN(64) and CNN(96) to denote them, respectively. The other one is the model averaging of different CNN architectures. We use AlexNet \cite{Alexnet} (much larger than LiuNet \cite{L-softmax}) and CIFAR10 Full (much smaller than LiuNet \cite{L-softmax}) architectures as an example, which have been provided in the standard Caffe \cite{caffe} library\footnote{\url{https://github.com/BVLC/caffe}.}. For comparison, all the architectures of these two kinds of model averaging strategies are equipped with the original softmax loss. Table \ref{ModelAverage} provides the experimental results of model averaging on CIFAR10 and CIFAR100 datasets, from which, we can see that the strategy of model averaging is beneficial to boost the classification performance. However, the training is time-consuming and the model size is large. Compared our weak classifiers ensemble (EM-Softmax) with these two kinds of model averaging, we can summarize that the accuracy of our EM-Softmax is general higher and our model size is much lower than the simple model averaging.

\begin{table}[t]
	\begin{center}
		\begin{tabular}{|c|c|c|c|c| }
		   \hline
		   Method & \multicolumn{2}{|c|}{CIFAR10} & \multicolumn{2}{|c|}{CIFAR100}  \\
		   \hline
           \hline
           LiuNet   & 15.1 & 90.95  & 15.7 & 67.26 \\
           Full     & 0.35 &  83.09 & 0.71 & 63.32  \\
           AlexNet  & 113.9 & 88.76 & 115.3 & 64.56 \\
           LiuNet+Full+Alex & 129.35 & 90.97 & 131.71 & 68.05 \\
           \hline
           LiuNet(48) & 9.3 & 90.63  & 9.4 & 66.08  \\
           LiuNet(64) & 15.1 & 90.95 & 15.3 & 66.70 \\
           LiuNet(96) & 30.9 & 91.16 & 31.0 & 67.26 \\
           LiuNet(48+64+96) & 55.4 & 91.99 & 55.7 & 70.01 \\
           \hline
           \multirow{3}*{}
           EM-Softmax  & 15.2 & \textbf{93.31}  & 31.1 & \textbf{72.74} \\
           \hline
        \end{tabular}
	 \end{center}		
\caption{The comparison of model size (MB) $|$ recognition accuracy rate (\%) of different ensemble strategies.}
\label{ModelAverage}
\end{table}

\subsection{EM-Softmax vs. Dropout}
Dropout is a popular way to train an ensemble and has been widely adopted in many works. The idea behind it is to train an ensemble of DNNs by randomly dropping the activations\footnote{Thus it cannot be applied to softmax classifier.} and averaging the results of the whole ensemble. The adopted architecture LiuNet \cite{L-softmax} contains the second last FC layer and is without the dropout. To validate the gain of our weak classifiers ensemble, we add the dropout technique to the second last fully-connected layer and conduct the experiments of Softmax, Softmax+Dropout and EM-Softmax+Dropout. The proportion of dropout is tuned in $\{0.3,0.5,0.7\}$ and the diversity hyperparameter $\lambda$ of our EM-Softmax is 0.1. Table \ref{Dropout} gives the experimental results of dropout on CIAFR10 and CIFAR100 datasets. From the numbers, we can see that the accuracy of our EM-Softmax is much higher than the Softmax+Dropout, which has shown the superiority of our weak classifier ensemble over the dropout strategy. Moreover, we emperically find that the improvement of dropout on both Softmax and EM-Softmax losses is not big with the adopted CNN architecture. To sum up, our weak classifiers ensemble is superior to the simple dropout strategy and can seamlessly incorporate with it.

\begin{table}[t]
	\begin{center}
		\begin{tabular}{|c|c|c| }
		   \hline
		   Method & \ CIFAR10 \  & \ CIFAR100  \\
		   \hline
           \hline
           Softmax          & 90.95  & 67.26 \\
           Softmax+Dropout  & 91.06  & 68.01\\
           \hline
           \multirow{3}*{}
           EM-Softmax  & 93.31  & 72.74 \\
           \hline
           \multirow{3}*{}
           EM-Softmax+Dropout  & \textbf{93.49}  & \textbf{72.85} \\
           \hline
        \end{tabular}
	 \end{center}		
\caption{Recognition accuracy rate (\%) vs. Dropout.}
\label{Dropout}
\end{table}

\subsection{Running Time}
We give the time cost vs. accuracy of EM-Softmax, Softmax and two kinds of model averaging on CIFAR10. From Table \ref{RunningTime}, the training time on 2 Titan X GPU is about 1.01h, 0.99h, 4.82h and 3.02h, respectively. The testing time on CPU (Intel Xeon E5-2660v0@2.20Ghz) is about 3.1m, 2.5m, 8.1m and 10m, respectively. While the corresponding accuracy is 93.31\%, 90.90\%, 90.97\% and 91.99\%, respectively. Considering time cost, model size and accuracy together, our weak classifiers ensemble EM-Softmax is a good candidate.

\begin{table}[t]
	\begin{center}
		\begin{tabular}{|c|c|c|c| }
		   \hline
           \multirow{2}*{Method}
		          & Training    & Test     & \multirow{2}*{Accuracy}  \\
                  & (GPU)       & (CPU)  &   \\
		   \hline
           \hline
           Softmax                  & 0.99h  & 2.5m  & 90.90 \\
           Softmax+Dropout          & 0.99h  & 2.5m  & 90.96 \\
           LiuNet+Full+AlexNet    & 4.82h  & 8.1m  & 90.97 \\
           LiuNet(48+64+96)  & 3.02h  & 10m   & 91.99 \\
           \hline
           EM-Softmax  & 1.01h  & 3.1m & \textbf{93.31} \\
           \hline
        \end{tabular}
	 \end{center}		
\caption{Recognition accuracy rate (\%) vs. Running time (h-hours, m-minutes) on \underline{CIFAR10} dataset.}
\label{RunningTime}
\end{table}

\section{Conclusion}
This paper has proposed a novel ensemble soft-margin softmax loss (\textit{i.e.}, EM-Softmax) for deep image classification. The proposed EM-Softmax loss benefits from two aspects. One is the designed soft-margin softmax function to make the learned CNN features to be discriminative. The other one is the ensemble weak classifiers to learn a strong classifier. Both of them can boost the performance. Extensive experiments on several benchmark datasets have demonstrated the advantages of our EM-Softmax loss over the baseline softmax loss and the state-of-the-art alternatives. The experiments have also shown that the proposed weak classifiers ensemble is generally better than model ensemble strategies (\textit{e.g.}, model averaging and dropout).


\section*{Acknowledgments}
This work was supported by National Key Research and Development Plan (Grant No.2016YFC0801002), the Chinese National Natural Science Foundation Projects $\#61473291$, $\#61572501$, $\#61502491$, $\#61572536$, the Science and Technology Development Fund of Macau (No.151/2017/A, 152/2017/A), and AuthenMetric R\&D Funds.

\bibliographystyle{named}
\bibliography{egbib}

\end{document}